# Supervised Machine Learning with a Novel Pointwise Density Estimator (arXiv:0710.5896)

Yen-Jen Oyang[1,*], Chien-Yu Chen[2], Darby Tien-Hao Chang[3], and Chih-Peng Wu[4]

[1] Graduate Institute of Biomedical Electronics and Bioinformatics, Department of Computer Science and Information Engineering, National Taiwan University, Taipei, 106, Taiwan, R.O.C. yjoyang@csie.ntu.edu.tw

[2] Department of Bio-Industrial Mechatronics Engineering, National Taiwan University, Taipei, Taiwan, R.O.C. chienyuchen@ntu.edu.tw

[3] Department of Electrical Engineering, National Cheng Kung University, Tainan, 70101, Taiwan, R.O.C. darby@ee.ncku.edu.tw

[4] Department of Computer Science and Information Engineering, National Taiwan University, Taipei, 106, Taiwan, R.O.C. hghung@mars.csie.ntu.edu.tw, chinuy@gmail.com

**Abstract.** This article proposes a novel density estimation based algorithm for carrying out supervised machine learning. The proposed algorithm features $O(n)$ time complexity for generating a classifier, where $n$ is the number of sampling instances in the training dataset. This feature is highly desirable in contemporary applications that involve large and still growing databases. In comparison with the kernel density estimation based approaches, the mathematical fundamental behind the proposed algorithm is not based on the assumption that the number of training instances approaches infinite. As a result, a classifier generated with the proposed algorithm may deliver higher prediction accuracy than the kernel density estimation based classifier in some cases.

**Keyterms:** density estimation, machine learning, data classification

## I. Introduction

In recent years, kernel density estimation has been exploited by computer scientists to model machine learning problems [1-3]. The kernel density estimation based approaches are of interest due to the low time complexity of either $O(n)$ or $O(n \log n)$ for generating an estimator [4], where $n$ is the number of sampling instances in the training dataset. This distinctive feature is highly desirable for handling contemporary databases that are already large and still growing. Nevertheless, because the theorems on which kernel density estimation is based are derived with the assumption that the number of training instances approaches infinite [4], machine learning algo-rithms based on kernel density estimation may suffer inferior prediction accuracy when the sampling density is not sufficiently high.

---

[*] To whom correspondences should be addressed. Tel:+886-2-33664888 ext. 431, Fax:+886-2-23688675.

This article proposes a novel pointwise density estimation algorithm and discusses how to exploit the proposed algorithm in data classification applications. The proposed algorithm features $O(n)$ time complexity for generating an estimator and its mathematical fundamental is not based on the assumption that the number of training instances approaches infinite.

## II. The Proposed Density Estimator

In this section, we will first elaborate the mathematical basis of the novel density estimator proposed in this article. Then, we will discuss how the proposed density estimator can be exploited in data classification applications.

Since we can always conduct a translation operation, without loss of generality, we can assume in the following discussion that it is the density at the origin of the coordinate system that is of concern. Furthermore, we can conduct a scaling operation so that all the sampling instances are within $\theta$ from the origin, where $0 \leq \theta < \frac{1}{2}$. Let $f_X(x_1, x_2, ..., x_m)$ denote the probability density function of the distribution of concern in an $m$-dimensional vector space. Assume that $f_X(x_1, x_2, ..., x_m)$ is analytic in the proximity of the origin and $f_X(x_1, x_2, ..., x_m) < \infty$ for all $(x_1, x_2, ..., x_m) \in \mathbf{R}^m$. Let $Z$ be the random variable that maps a sampling instance $s_i$ taken from the distribution governed by $f_X$ to $\|\mathbf{s}_i\|^m$, where $\|\mathbf{s}_i\|$ is the distance between the origin and $s_i$. Accordingly, we have the distribution function $F_Z(z)$ of $Z$ equal to

$$\iint...\int_{x_1^2+x_2^2+...+x_m^2 \leq z^{2/m}} f_X(x_1, x_2, ..., x_m) dx_1 dx_2 ... dx_m$$

for $z \geq 0$ and $F_Z(z) = 0$ for $z < 0$.

**Theorem 1:** Let $f_Z(z) = \lim_{\substack{\varepsilon \to 0 \\ \varepsilon > 0}} \frac{F_Z(z+\varepsilon) - F_Z(z)}{\varepsilon}$ for $z \geq 0$. Then, we have $f_Z(0) = \frac{\pi^{m/2}}{\Gamma(m/2+1)} f_X(\mathbf{0})$, where $\Gamma(\cdot)$ is the gamma function [5].

**Proof:**
   Since $F_Z(z) = 0$ we have

$$\lim_{\substack{\varepsilon \to 0 \\ \varepsilon > 0}} \frac{F_Z(\varepsilon) - F_Z(0)}{\varepsilon}$$

$$= \lim_{\substack{\varepsilon \to 0 \\ \varepsilon > 0}} \frac{\int\!\!\int\cdots\!\int_{x_1^2+x_2^2+\ldots+x_m^2 \leq \varepsilon^{2/m}} f_X(x_1, x_2, \ldots, x_m) dx_1 dx_2 \ldots dx_m}{\varepsilon}$$

By the Taylor expansion,
$$f_X(x_1, x_2, \ldots, x_m)$$
$$= f_X(\mathbf{0}) + \frac{\partial f_X(\mathbf{0})}{\partial x_1} x_1 + \ldots + \frac{\partial f_X(\mathbf{0})}{\partial x_m} x_m + \text{high - order term s.}$$

Furthermore, in region where $x_1^2 + x_2^2 + \ldots + x_m^2 \leq \varepsilon^{2/m}$, we have $x_1 \to 0$, $x_2 \to 0$, ..., $x_m \to 0$ as $\varepsilon \to 0$. Therefore,

$$f_Z(0) = \lim_{\substack{\varepsilon \to 0 \\ \varepsilon > 0}} f_X(\mathbf{0}) \cdot \frac{(\sqrt[m]{\varepsilon})^m \pi^{m/2}}{\Gamma(m/2+1)} \cdot \frac{1}{\varepsilon}$$

$$= \frac{\pi^{m/2}}{\Gamma(m/2+1)} \cdot f_X(\mathbf{0}),$$

where $\frac{(\sqrt[m]{\varepsilon})^m \pi^{m/2}}{\Gamma(m/2+1)}$ is the volume of a sphere in an *m*-dimensional vector space with radius $= \sqrt[m]{\varepsilon}$.

□

Theorem 1 implies that we can obtain an estimate of $f_X(\mathbf{0})$ by first obtaining an estimate of $f_Z(0)$. Assume that $g(r)$ defined by

$$g(r) = \frac{1}{\beta \cdot r^{m-1}} \cdot \frac{d}{dr}\left[\int\!\!\int\cdots\!\int_{x_1^2+x_2^2+\ldots+x_m^2 \leq r^2} f_X(x_1, x_2, \ldots, x_m) dx_1 dx_2 \ldots dx_m\right] \text{ for } r > 0 \text{ and}$$

$$g(0) = \lim_{\substack{r \to 0 \\ r > 0}} \frac{1}{\beta \cdot r^{m-1}} \cdot \frac{d}{dr}\left[\int\!\!\int\cdots\!\int_{x_1^2+x_2^2+\ldots+x_m^2 \leq r^2} f_X(x_1, x_2, \ldots, x_m) dx_1 dx_2 \ldots dx_m\right]$$

is analytic with all orders of the right-sided derivatives at 0 for $r \in [0, \theta]$, where $\beta = \frac{2\pi^{m/2}}{\Gamma(m/2)}$ and $\frac{2\pi^{m/2}}{\Gamma(m/2)} r^{m-1}$ is the surface area of a sphere with radius *r* in an *m*- dimensional vector space. Accordingly, further assume that $g(r) = \sum_{j=0}^{\infty} \lambda_j r^j$ for $r \in [0, \theta]$. Then, for $z \in [0, \theta^m]$,

$$F_Z(z) = \int_0^{z^{1/m}} \beta \cdot r^{m-1} g(r) dr$$

$$= \int_0^{z^{1/m}} [\beta \cdot \sum_{j=0}^{\infty} \lambda_j r^{m+j-1}] dr$$

$$= \beta \cdot \sum_{j=0}^{\infty} \frac{\lambda_j z^{1+j/m}}{m+j}.$$

Accordingly, $f_Z(z) = \frac{\beta}{m} \sum_{j=0}^{\infty} \lambda_j z^{j/m}$ for $z \in [0, \theta^m]$.

In practical implementation, for computing $f_Z(z)$ with $z \in [0, \theta^m]$, we only need to consider a limited number of terms in $\sum_{j=0}^{\infty} \lambda_j z^{j/m}$. Assume that it is determined that only $\lambda_0, \lambda_1, \lambda_2, ..., \lambda_q$ are of concern, we then can figure out an optimal combinations of $\lambda_0, \lambda_1, \lambda_2, ..., \lambda_q$ based on the maximum-likelihood approach. That is, we can employ an optimization algorithm to figure out the values of $\lambda_0, \lambda_1, \lambda_2, ..., \lambda_q$ that maximize

$$\sum_{i=1}^{n} \ln\left(\sum_{j=0}^{q} \lambda_j \|s_i\|^j\right) \text{ subject to } \int_0^{\theta^m} f_Z(z) dz = 1,$$

where $\{s_1, s_2, ..., s_n\}$ is the set of training instances. Once we obtain an optimal combination of the values of $\lambda_0, \lambda_1, \lambda_2, ..., \lambda_q$, then $\frac{\beta}{m} \lambda_0$ is the estimate of $f_Z(0)$ and $\frac{\beta \cdot \Gamma(m/2+1)}{m \cdot \pi^{m/2}} \cdot \lambda_0 = \lambda_0$ is an estimate of $f_X(\mathbf{0})$.

The proposed density estimator is referred to as the super-radius based density estimator (SRDE). It is so named because random variable $Z$ maps a sampling instance $s_i$ taken from the distribution governed by $f_X$ to $\|\mathbf{s}_i\|^m$ and $\|\mathbf{s}_i\|^m$ is referred to as the super-radius of $s_i$ in this article. For data classification applications, a query instance located at $\mathbf{v}$ is predicted to belong to the class that gives the maximum value among the likelihood functions defined in the following:

$$L_j(\mathbf{v}) = \frac{|S_j| \cdot \hat{f}_j^*(\mathbf{v})}{\sum_h |S_h| \cdot \hat{f}_h^*(\mathbf{v})},$$

where $|S_j|$ is the number of class-$j$ training instances and $\hat{f}_j^*(\mathbf{v})$ is the pointwise estimate of the density of class-$j$ training instances at $\mathbf{v}$. In our current implementation, aiming to improve the execution time of the classifier, we include only a limited number, denoted by $k$, of the nearest class-$j$ training instances of $\mathbf{v}$ in computing $\hat{f}_j^*(\mathbf{v})$.

The training time of the SRDE based classifier is in the order of $O(n)$ because the only thing to be carried out is to scale the training dataset so that all the sampling instances are within $\theta$ from the origin and the distance between any two training instances is less than 1.

## Remarks

The results presented in this article imply that the main result obtained in our previous article [6] is not generally applicable. The main result obtained in our previous article is based on the assumption that $f_Z(z)$ is analytic. However, for $f_Z(z)$ to be analytic, many coefficients in $\sum_{j=0}^{\infty} \lambda_j z^{j/m}$ must be zero and this is a very strict condition. What we have proposed in this article is an approach that is generally applicable for exploiting Theorem 1.

The results presented in this article can be generalized to make pointwise estimation of the value of a multidimensional function. Provided that $h(x_1, x_2, ..., x_m)$ is a function in an $m$-dimensional vector space that is analytic in the proximity of the origin and $h(x_1, x_2, ..., x_m) < \infty$ for all $(x_1, x_2, ..., x_m) \in \mathbf{R}^m$, then define

$$H(z) = \iint ... \int_{x_1^2 + x_2^2 + ... + x_m^2 \leq z^{2/m}} h(x_1, x_2, ..., x_m) dx_1 dx_2 ... dx_m$$

for $z \geq 0$ and $H(z) = 0$ for $z < 0$. By following pretty much the same procedure as presented in the previous section, we can obtain that $\lambda_0$ is an estimate of $h(\mathbf{0})$. The only difference is that for pointwise estimation of function value we may want to figure out the values of $\lambda_0, \lambda_1, \lambda_2, ..., \lambda_q$ that minimize

$$\left| \sum_{i=1}^{n} \left\{ h(\mathbf{s}_i) - \left( \sum_{j=0}^{q} \lambda_j \|\mathbf{s}_i\|^j \right) \right\} \right|,$$

where $\{h(\mathbf{s}_1), h(\mathbf{s}_2), ..., h(\mathbf{s}_n)\}$ is the set of training instances.

## Contributions of Authors

YJO initiated this study, proposed the SRDE, and established its mathematical foundation. CYC, DTHC, and CPW jointly implemented the software employed to verify the correctness of the mathematical fundamentals.

## Acknowledgement

The authors greatly appreciate the valuable comments and suggestions provided by Prof. Hung Chen of National Taiwan University, Prof. Henry Horng-Shing Lu of National Chiao-Tung University, and Prof. Yuan-Chin Chang of Academia Sinica. This research has been supported by the National Science Council of R.O.C. under the contracts NSC 95-3114-P-002-005-Y and NSC 96-2627-B-002-003.